\documentclass[times,10pt,twocolumn]{article}
\usepackage{latex8,times}
\usepackage{amsmath,amssymb}
\usepackage{pstricks,pst-node}

\newcommand{\copyrightnotice}[1]
	{#1}	%%  print
\newcommand{\bfemph}[1]		{{\bfseries\emph{#1}}}

\newenvironment{example}	{\paragraph*{Example.}}{}
\newcommand{\qedex}		{\hfill$\Box$}

\newcommand{\eclipse}		{\textup{ECL\textsuperscript{\textit{i}}PS%
				\textsuperscript{\textit{e}}}}
\newcommand{\RCC}		{\mathsf{RCC8}}
\newcommand{\PS}		{\mathcal{PS}}

\newcommand{\mi}[1]		{\mathit{#1}}
\newcommand{\ms}[1]		{\mathsf{#1}}

\newcommand{\mt}[1]		{\mathtt{#1}}

\newcommand{\lAnd}		{\bigwedge}
\newcommand{\lOr}		{\bigvee}

\newcommand{\setUnion}		{\bigcup}
\newcommand{\limplies}		{\rightarrow}

\newcommand{\tuple}[1]		{\langle #1 \rangle}
\newcommand{\rangebopd}[5]	{#1_#2#4 #5 #4#1_#3}
\newcommand{\rangeb}[3]		{\rangebopd{#1}{#2}{#3},{\ldots}}

\newcommand{\rangecbop}[4]	{\rangebopd{#1}{#2}{#3}{#4}{\cdots}}

\renewcommand{\leq}		{\leqslant}
\renewcommand{\geq}		{\geqslant}

\renewcommand{\emptyset}	{\varnothing}

\newcommand{\setnewlength}[2]	{\newlength{#1}\setlength{#1}{#2}}

\newcommand{\F}			{\mathcal{F}}
\newcommand{\C}			{\mathcal{C}}

\newcommand{\I}			{\mathcal{I}}
\newcommand{\Q}			{\mathcal{Q}}
\newcommand{\OO}		{\mathcal{O}}
\newcommand{\PP}		{\mathcal{P}}
\newcommand{\SSS}		{\mathcal{S}}

\newcommand{\ie}		{i.\,e.}
\newcommand{\eg}		{e.\,g.}

\newcommand{\ra}		{\rightarrow}
\newcommand{\raa}		{\ \rightarrow\ }
\newcommand{\so}[1]		{\mbox{\textsl{#1}}}

\newcommand{\disjoint}		{\mathsf{disjoint}}
\newcommand{\meet}		{\mathsf{meet}}
\newcommand{\overlap}		{\mathsf{overlap}}
\newcommand{\equal}		{\mathsf{equal}}
\newcommand{\covers}		{\mathsf{covers}}
\newcommand{\coveredby}		{\mathsf{coveredby}}
\newcommand{\contains}		{\mathsf{contains}}
\newcommand{\inside}		{\mathsf{inside}}

\newcommand{\north}		{\mathsf{N}}
\newcommand{\northwest}		{\mathsf{NW}}
\newcommand{\northeast}		{\mathsf{NE}}
\newcommand{\south}		{\mathsf{S}}
\newcommand{\southwest}		{\mathsf{SW}}

\newcommand{\east}		{\mathsf{E}}
\newcommand{\west}		{\mathsf{W}}

\newcommand{\samepoint}		{\mathsf{EQ}}

\newcommand{\comp}		{\mathsf{comp}}
\newcommand{\conv}		{\mathsf{conv}}

\newcommand{\always}		{{\psset{unit=0.07em,linewidth=0.5}\begin{pspicture}(11,10)\psframe[linearc=0.01](1,0)(10,10)\end{pspicture}}}
\newcommand{\eventually}	{{\psset{unit=0.07em,linewidth=0.5}\begin{pspicture}(11,10.2)\psdiamond[linearc=0.01](5,5)(5.1,5.1)\end{pspicture}}}
\newcommand{\nexttime}		{{\psset{unit=0.07em,linewidth=0.5}\begin{pspicture}(11,10)\pscircle(5,5){5}\end{pspicture}}}
\newcommand{\until}		{\mathbin{\mathsf{U}}}
\newcommand{\since}		{\mathbin{\mathsf{S}}}

\begin{document}

\title{Constraint-Based Qualitative Simulation}

\author{Krzysztof R. Apt\\
	National University of Singapore, Singapore, and\\
	CWI and UvA, Amsterdam, The Netherlands\\
	apt@comp.nus.edu.sg
	\and
	Sebastian Brand\\
	National University of Singapore, Singapore\\
	brand@comp.nus.edu.sg}

\maketitle
\thispagestyle{empty}

\begin{abstract}
  We consider qualitative simulation involving a finite set
  of qualitative relations in presence of complete
  knowledge about their interrelationship. We show how it
  can be naturally captured by means of constraints
  expressed in temporal logic and constraint satisfaction
  problems. The constraints relate at each stage the `past'
  of a simulation with its `future'.  The benefit of this
  approach is that it readily leads to an implementation
  based on constraint technology that can be used to
  generate simulations and to answer queries about them.
\end{abstract}

\copyrightnotice{
\footnotetext[0]{
\copyright 2005 IEEE. Personal use of this material is permitted.
However, permission to reprint/republish this material for advertising or
promotional purposes or for creating new collective works for resale
or redistribution to servers or lists, or to reuse any copyrighted
component of this work in other works must be obtained from the IEEE.
}
}

\section{Introduction}

\bfemph{Qualitative reasoning} was introduced in AI
to abstract from numeric quantities, such as the precise
time of an event or the location or trajectory of an
object in space, and to reason instead on the level
of appropriate abstractions.
Two different forms of qualitative reasoning
were studied in the literature.
The first one is concerned with reasoning about continuous
change in physical systems, monitoring streams of observations and
simulating behaviours, to name a few applications.
The main techniques used are qualitative differential equations,
constraint propagation and discrete state graphs.
For a thorough introduction see \cite{Kui94}.

The second form of qualitative aims at reasoning about contingencies
such as time, space, shape, size, directions, through an
abstraction of the quantitative information into a finite
set of qualitative relations. One then relies on complete
knowledge about the interrelationship of these qualitative
relations.  This approach is exemplified by temporal
reasoning due to \cite{All83},
spatial reasoning introduced in \cite{Ege91} and \cite{randell:1992:spatial},
reasoning about cardinal directions
(such as North, Northwest),
see, \eg, \cite{Lig98}, etc.
For a recent overview of this approach
to spatial reasoning, see \cite{CH01}.

\bfemph{Qualitative simulation} deals with reasoning
about possible evolutions in time of models capturing qualitative information.
One assumes that time is discrete and that only
changes adhering to some desired format occur at each stage.
\cite{Kui01} discusses qualitative simulation in the first framework,
while \bfemph{qualitative spatial simulation} is considered in \cite{CCR92a}.

Our aim here is to show how qualitative simulation in the
second approach to qualitative reasoning
(exemplified by qualitative temporal and spatial reasoning)
can be naturally captured by
means of temporal logic and constraint satisfaction problems.
The resulting framework allows us to concisely
describe various complex forms of behaviour,
such as a simulation of a naval navigation problem
or a solution to a version of a piano movers problem.
The domain knowledge is formulated using a variant of
linear temporal logic with both past and future temporal
operators.
Such temporal formulas are then translated into constraints.

The usual constraint-oriented representation
of the second approach to qualitative reasoning
is based on
modelling qualitative relations as constraints.
See, for example, \cite{escrig:1998:qualitative}
for an application of this modelling approach.
In contrast, we
represent qualitative relations as variables.
This way of modelling has important advantages.
In particular, it is more declarative since
model and solver are kept separate;
see the study of the relation variable model
in \cite{brand:2004:relation}.
In our case it allows us to express all
domain knowledge on the same conceptual level,
namely as constraints on the relation variables.
Standard techniques of constraint programming can then be used
to generate the simulations and to answer queries about them.

To support this claim, we implemented this approach
in the generic constraint programming system \eclipse{}
\cite{wallace:1997:eclipse}
and discuss here several case studies.

%=====================================================================

\section{Simulation Constraints}\label{sec:constraints}

\subsection{Constraint Satisfaction Problems}\label{sec:csp}

We begin by briefly introducing Constraint Programming.
Consider a sequence  $X = \rangeb x1m$ of
variables with respective domains $\rangeb D1m$. By
a \bfemph{constraint} $C$ on $X$, written $C(X)$,
we mean a subset of $\rangecbop D1m{\times}$.
A \bfemph{constraint satisfaction problem (CSP)} consists of a finite sequence
of variables $X$ with respective domains
and a finite set $\C$ of constraints, each on a subsequence of $X$\@.
A \bfemph{solution} to a CSP
is an assignment to its variables respecting their domains and constraints.

We study here CSPs with finite domains.
They can be solved by a \bfemph{top-down search}
interleaved with \bfemph{constraint propagation}.
The top-down search is determined by a
\bfemph{branching} strategy that controls the splitting
of a given CSP into two or more CSPs,
the `union' of which is equivalent to (\ie, has the
same solutions as) the initial CSP.
In turn, constraint propagation transforms
a given CSP into one that is equivalent but \emph{simpler}.
We use here heuristics-controlled
\bfemph{domain partitioning} as the branching strategy
and \bfemph{hyper-arc consistency} of \cite{MM88}
as the constraint propagation.
Hyper-arc consistency is enforced by removing
from each variable domain
the elements not used in a constraint.

\subsection{Intra-state Constraints}\label{sec:intrastate}

To describe qualitative simulations formally,
we define first intra-state and inter-state constraints.
A qualitative simulation corresponds then to a
CSP consisting of \emph{stages} that
all satisfy the intra-state constraints.
Moreover, this CSP satisfies the inter-state constraints
that link the variables appearing in various stages.

For presentational reasons, we restrict ourselves here to
binary qualitative relations (\eg, topology, relative size).
This is no fundamental limitation;
our approach extends directly to higher-arity relations
(\eg, ternary orientation).

\smallskip
We assume that we have at our disposal
\begin{itemize}
\item	a finite \bfemph{set of qualitative relations} $\Q$,
	with a special element
	denoting the relation of an object to itself;
\item	consistency conditions on $\Q$-scenarios;
	we assume the usual case that they can be
	expressed as relations over $\Q$,
	specifically as
	a binary \bfemph{converse} relation $\conv$
	and a ternary \bfemph{composition} relation $\comp$,
\item	a \bfemph{conceptual neighbourhood} relation
	%$\neighbour$
	between the elements of $\Q$
	that describes which \emph{atomic} changes
	in the qualitative relations are admissible.
\end{itemize}

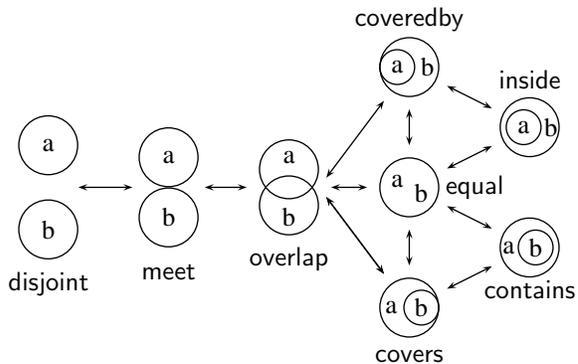
\begin{figure}
\centering
\psset{unit=8mm,arrows=<->, nodesep=0.2, linewidth=0.5pt}
\begin{pspicture}(0,0)(10,6)

\pscircle(1,3.7){0.5}\rput(1,3.7){a}
\pscircle(1,2.3){0.5}\rput(1,2.3){b}
\pnode(1.3,3){disjoint}
\rput(1,1.4){$\disjoint$}

\pnode(2.6,3){meet1}
\pscircle(3,3.5){0.5}\rput(3,3.5){a}
\pscircle(3,2.5){0.5}\rput(3,2.5){b}
\pnode(3.4,3){meet2}
\rput(3,1.6){$\meet$}

\pnode(4.5,3){overlap1}
\pscircle(5,3.3){0.5}\rput(5,3.5){a}
\pscircle(5,2.7){0.5}\rput(5,2.5){b}
\pnode(5.5,3){overlap2}
\rput(5,1.8){$\overlap$}

\cnode(7,3){0.5}{equal}\rput(6.8,3.1){a}
\rput[l](7.6,3){$\equal$}\rput(7.2,2.9){b}

\cnode(7,5){0.5}{coveredby}\rput(6.8,5){a}
\pscircle(6.8,5){0.3}\rput(7.3,5){b}
\rput(7,5.8){$\coveredby$}

\cnode(7,1){0.5}{covers}\rput(6.7,1){a}
\pscircle(7.2,1){0.3}\rput(7.2,1){b}
\rput(7,0.2){$\covers$}

\cnode(9,4){0.5}{inside}\rput(8.9,4){a}
\pscircle(8.9,4){0.3}\rput(9.35,4){b}
\rput(9,4.8){$\inside$}

\cnode(9,2){0.5}{contains}\rput(8.65,2){a}
\pscircle(9.1,2){0.3}\rput(9.1,2){b}
\rput(9,1.3){$\contains$}

\ncline{disjoint}{meet1}
\ncline{meet2}{overlap1}
\ncline{overlap2}{covers}
\ncline{overlap2}{equal}
\ncline{overlap2}{covers}
\ncline{overlap2}{coveredby}
\ncline{equal}{inside}
\ncline{equal}{contains}
\ncline{covers}{contains}
\ncline{coveredby}{inside}
\ncline{equal}{covers}
\ncline{equal}{coveredby}
\end{pspicture}
\caption{The eight $\RCC$ relations}
\label{fig:evolution}
\end{figure}

\begin{example}\label{ex:rcc8}
Take the qualitative spatial reasoning with topology
introduced in \cite{Ege91} and \cite{randell:1992:spatial}.
The set of qualitative relations is the set $\RCC$, \ie,
\[\begin{array}{@{}l@{}l}
	\Q =  \{ & \disjoint,\meet,\equal,\covers,\\
		& \coveredby,\contains,\inside,\overlap \};
\end{array}\]
see Fig.~\ref{fig:evolution}, which also shows
the neighbourhood relation between these relations.
\qedex
\end{example}

We fix now a sequence $\OO$ of objects of interest.
By a \bfemph{qualitative array} we mean a
two-dimensional array $Q$ on $\OO \times \OO$ such that
\begin{itemize}
\item for each pair of objects $\mt{A, B} \in \OO$,
	the expression $Q[\mt{A,B}]$ is a variable
	denoting the (basic) relation between $\mt{A, B}$.
	So its initial domain is a subset of $\Q$.

\item the consistency conditions hold on $Q$, so
	for each triple of objects $\mt{A, B, C}$
	the following \bfemph{intra-state constraints} are satisfied:
	%\begin{itemize}
	%\item \bfemph{reflexivity}:	$Q[\mt{A,A}] = \equal$,
	%\item \bfemph{converse}:	$\conv(Q[\mt{A,B}], Q[\mt{B,A}])$,
	%\item \bfemph{composition}:	$\comp(Q[\mt{A,B}], Q[\mt{B,C}], Q[\mt{A,C}])$.
	%\end{itemize}

	\smallskip
	\begin{tabular}{@{\quad}ll}
	\bfemph{reflexivity}:	& $Q[\mt{A,A}] = \equal$,\\[1ex]
	\bfemph{converse}:	& $\conv(Q[\mt{A,B}], Q[\mt{B,A}])$,\\[1ex]
	\bfemph{composition}:	& $\comp(Q[\mt{A,B}], Q[\mt{B,C}], Q[\mt{A,C}])$.
	\end{tabular}
\end{itemize}

Each qualitative array determines a unique CSP.
Its variables are $Q[\mt{A,B}]$, with $\mt{A}$ and $\mt{B}$
ranging over the sequence of the assumed objects $\OO$.
The domains of these variables are appropriate subsets of~$\Q$.
An instantiation of the variables to elements of $Q$
corresponds to a consistent $Q$-scenario.

In what follows we represent each stage $t$ of a
simulation by a CSP $\PP_t$ uniquely determined
by a qualitative array $Q_t$.
Here $t$ is a variable ranging over the set of
natural numbers that represents discrete time.
Instead of $Q_t[\mt{A,B}]$ we also write $Q[\mt{A,B},t]$,
reflecting that, in fact, we deal with a single \emph{ternary} array.

%=====================================================================

\subsection{Inter-state Constraints}\label{sec:interstate}

To describe the inter-state constraints, we use as
\bfemph{atomic formulas} statements of the form
\[
	Q[\mt{A,B}] \mathbin{?} q
\]
where $? \in \{=, \neq\}$ and $q \in \Q$, or `$\ms{true}$',
and employ a temporal logic with four \bfemph{temporal operators},
\[\begin{array}{ll@{\quad}ll}
 \eventually	&\text{(eventually)},&
 \nexttime	&\text{(next time)},\\
 \always	&\text{(from now on)},&
 \until		&\text{(until)},
\end{array}\]
and their `past' counterparts,
$\nexttime^{-1}$, $\eventually^{-1}$,
$\always^{-1}$, and $\since$ (since).
While it is known that past time operators
can be eliminated, their use
results in more succinct
(and in our case more intuitive) specifications;
see, \eg, \cite{LMS02}.

\bfemph{Inter-state constraints} are formulas
that have the form $\phi \ra \nexttime \psi$.
Both $\phi$ and $\psi$ are built out of atomic formulas using
propositional connectives,
but $\phi$ contains only past time temporal operators
and $\psi$ uses only future time operators.

Intuitively, at each time instance $t$,
each inter-state constraint $\phi \ra \nexttime \psi$
links the `past' CSP
$\setUnion_{i = 0}^{t} \PP_i$ with the `future' CSP
$\setUnion_{i = t + 1}^{t_{\max}} \PP_i$.
% where $t_{\max}$ is the fixed maximum
% length of the simulation.
So we interpret $\phi$ in the interval $[0..t]$,
and $\psi$ in the interval $[t+1 \mathbin{..} t_{\max}]$.

We now explain the meaning of a past or future
temporal formula $\phi$ with respect to the
underlying qualitative array $Q$
in an interval $[s..t]$,
for which we stipulate $s \leq t$.
We write $\models_{[s..t]} \phi$
to express that $\phi$ holds in the interval $[s..t]$.

%---------------------------------------------------------------------

\paragraph{Propositional connectives.}

These are defined as expected,
in particular independently of the `past' or `future'
aspect of the formula. For example,
\[\begin{array}{l@{\hspace{1.5em}}l@{\hspace{1.5em}}l}
	\models_{[s..t]} \lnot \phi
	&\text{if not}&
	\models_{[s..t]} \phi, 
	\\
	\models_{[s..t]} \phi_1 \lor \phi_2
	&\text{if}&
	\models_{[s..t]} \phi_1 \text{ or } \models_{[s..t]} \phi_2.
	\end{array}\]
Conjunction $\phi_1 \land \phi_2$
and implication $\phi_1 \limplies \phi_2$
are defined analogously.

% abbreviates $\lnot\phi_1 \lor \phi_2$.

%---------------------------------------------------------------------

\paragraph{Future formulas.}

Intuitively, the evaluation starts
at the lower bound of the time interval
and moves only forward in time.
\[\begin{array}{l@{\hspace{1.5em}}l@{\hspace{1.5em}}l}
	\models_{[s..t]} Q[A,B] \mathbin{?} c
		&\text{if}&
		Q[A,B,s] \mathbin{?} c
		\\&&
		\text{where } ? \in \{=,\neq\};
		\\[1ex]
	\models_{[s..t]} \nexttime \phi
		&\text{if}&
		\models_{[r..t]} \phi
		\\&&
		\text{and }r = s+1 \text{ and } r \leq t;
		\\
	\models_{[s..t]} \always \phi
		&\text{if}&
		\models_{[r..t]} \phi
		\text{ for all } r \in [s..t];
		\\
	\models_{[s..t]} \eventually \phi
		&\text{if}&
		\models_{[r..t]} \phi
		\text{ for some } r \in [s..t];
		\\
	\models_{[s..t]} \chi \until \phi
		&\text{if}&
		\models_{[r..t]} \phi
		\text{ for some } r \in [s..t]
		\\&&
		\hspace*{-2.5em}\text{and }
		\models_{[u..t]} \chi
		\text{ for all } u \in [s \mathbin{..} r-1].
\end{array}\]

%---------------------------------------------------------------------

\paragraph{Past formulas.}

Here the evaluation starts
at the upper bound and moves backward.
\[\begin{array}{l@{\hspace{1.5em}}l@{\hspace{1.5em}}l}
	\models_{[s..t]} Q[A,B] \mathbin{?} c
		&\text{if}&
		Q[A,B,t] \mathbin{?} c
		\\&&
		\text{where } ? \in \{=,\neq\};
		\\[1ex]
	\models_{[s..t]} \nexttime^{-1} \phi
		&\text{if}&
		\models_{[s..r]} \phi
		\\&&\text{and }
		r = t-1 \text{ and } s \leq r;
		\\
	\models_{[s..t]} \always^{-1} \phi
		&\text{if}&
		\models_{[s..r]} \phi
		\text{ for all } r \in [s..t];
		\\
	\models_{[s..t]} \eventually^{-1} \phi
		&\text{if}&
		\models_{[s..r]} \phi
		\text{ for some }
		r \in [s..t];
		\\
	\models_{[s..t]} \chi \since \phi
		&\text{if}&
		\models_{[s..r]} \phi
		\text{ for some } r \in [s..t]
		\\&&
		\hspace*{-2.5em}\text{and }
		\models_{[u..t]} \chi
		\text{ for all }
		u \in [r+1 \mathbin{..} t].
\end{array}\]

Furthermore,
we write $Q[\mt{A,B}] \in \{\rangeb q1k\}$ 
as an abbreviation of
$(Q[\mt{A,B}] = q_1) \lor \ldots \lor (Q[\mt{A,B}] = q_k)$.
The meaning of $Q[\mt{A,B}] \notin \{\rangeb q1k\}$
is analogous.

The bounded quantification
$\exists \mt{A} \in \{\rangeb o1k\}.\, \phi(\mt{A})$
represents the disjunction 
$\phi(o_1) \lor \ldots \lor \phi(o_k)$.
Universal quantification
$\forall \mt{A} \in \{\rangeb o1k\}.\,\phi(\mt{A})$
is interpreted analogously.
As usual, $\mt{A}$ in $\phi(\mt{A})$ denotes a placeholder
(free variable), and $\phi(o_i)$ is obtained
by replacing $\mt{A}$ in all its occurrences by $o_i$.

%---------------------------------------------------------------------

\subsection{An Example: Navigation}

\newcommand{\sh}	{\so{ship}}
\newcommand{\bu}	{\so{buoy}}

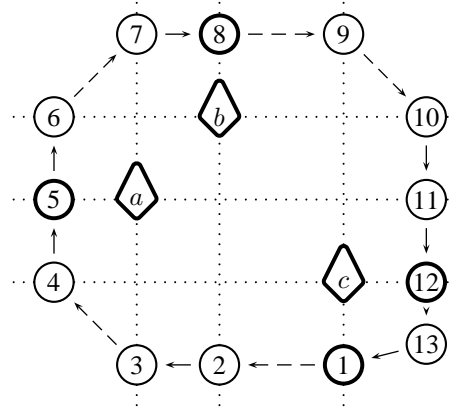
\begin{figure}
\centering
\psset{unit=5.5mm}
\begin{pspicture}(0,0)(11,10)
{\psset{linestyle=dotted}
\psline(5,0)(5,10)
\psline(3,0)(3,10)
\psline(8,0)(8,10)
\psline(0,7)(11,7)
\psline(0,5)(11,5)
\psline(0,3)(11,3)
}%

{\psset{fillstyle=solid,linearc=0.1,linewidth=1.5pt}
\pspolygon(2.5,5)(3,4.5)(3.5,5)(3,6)\rput(3,5){$a$}
\pspolygon(4.5,7)(5,6.5)(5.5,7)(5,8)\rput(5,7){$b$}
\pspolygon(7.5,3)(8,2.5)(8.5,3)(8,4)\rput(8,3){$c$}
}%

{\psset{radius=0.5,fillstyle=solid}
\Cnode[linewidth=1.5pt](8,1){p1}\rput(8,1){1}
\Cnode(5,1){p2}\rput(5,1){2}
\Cnode(3,1){p3}\rput(3,1){3}
\Cnode(1,3){p4}\rput(1,3){4}
\Cnode[linewidth=1.5pt](1,5){p5}\rput(1,5){5}
\Cnode(1,7){p6}\rput(1,7){6}
\Cnode(3,9){p7}\rput(3,9){7}
\Cnode[linewidth=1.5pt](5,9){p8}\rput(5,9){8}
\Cnode(8,9){p9}\rput(8,9){9}
\Cnode(10,7){p10}\rput(10,7){10}
\Cnode(10,5){p11}\rput(10,5){11}
\Cnode[linewidth=1.5pt](10,3){p12}\rput(10,3){12}
\Cnode(10,1.5){p13}\rput(10,1.5){13}
}%

{\psset{linestyle=dashed,arrows=->,linewidth=0.5pt,nodesep=3pt}
\ncline{p1}{p2}
\ncline{p2}{p3}
\ncline{p3}{p4}
\ncline{p4}{p5}
\ncline{p5}{p6}
\ncline{p6}{p7}
\ncline{p7}{p8}
\ncline{p8}{p9}
\ncline{p9}{p10}
\ncline{p10}{p11}
\ncline{p11}{p12}
\ncline{p12}{p13}
\ncline{p13}{p1}
}%
\end{pspicture}
\caption{Navigation path}
\label{fig:navigation}
\end{figure}

A ship navigates around three buoys along a specified course.
The position of the buoys is fixed;
see Fig.~\ref{fig:navigation}.
We reason qualitatively about the cardinal directions
\[
	\Q = \{ \north, \northeast, \ldots, \west, \northwest,
		\ \samepoint\}
\]
with the obvious meaning
($\samepoint$ is the identity relation).
Ligozat \cite{Lig98} provides the composition table for this form of
qualitative reasoning and shows that it captures consistency.

The buoy positions are given by the following
global intra-state constraints:
\begin{align*}
Q[\bu_a, \bu_c] &= \northwest,\\
Q[\bu_a, \bu_b] &= \southwest,\\
Q[\bu_b, \bu_c] &= \northwest.
\end{align*}

All objects occupy different positions:
\[
	\forall \mt{A},\mt{B} \in \mathcal{O}.\
	\mt{A} \neq \mt{B} \raa
	Q[\mt{A},\mt{B}] \neq \samepoint.
\]

The initial position of the ship is south of buoy $c$,
so we have $Q[\sh, \bu_c] = \south$.
The ship is required to follow a path around the buoys.
In Fig.~\ref{fig:navigation},
the positions required to be visited
are marked with bold circles.
We stipulate
\[\begin{array}{l}
	\eventually (Q[\sh, \bu_a] = \west
	\ \land{} \\[1ex]
	\hspace{1.2em}\eventually (Q[\sh, \bu_b] = \north
	\ \land{} \\[1ex]
	\hspace{2.4em}\eventually (Q[\sh, \bu_c] = \east
	\ \land{} \\[1ex]
	\hspace{3.6em}\eventually (Q[\sh, \bu_c] = \south
	\ ))))
\end{array}\]
to hold in the interval $[0 \mathbin{..} t_{\max}]$.

A tour of 13 steps exists (and is found by our program);
it is indicated in Fig.~\ref{fig:navigation}.

%=====================================================================

\section{Temporal Formulas as Constraints}\label{sec:translations}

\newcommand{\cons}	{\mathit{cons}}
\newcommand{\consc}[3]	{\mathit{cons}([#2..#3], #1)}
\newcommand{\conscd}[4]	{\mathit{cons}^{#1}([#3..#4], #2)}
\newcommand{\consr}[4]	{\consc{#1}{#2}{#3} \equiv #4}
\newcommand{\consrd}[5]	{\conscd{#1}{#2}{#3}{#4} \equiv #5}

We explain now how a temporal formula
(an inter-state constraint)
is imposed on the sequence of CSPs representing the
spatial arrays at consecutive times.
Such a formula is reduced
to a sequence of constraints
by eliminating the temporal operators.
We provide two alternative translations.
The first simply unfolds the temporal operators
into primitive constraints,
while the second
retains more structure and avoids duplication of subformulas by
relying on \emph{array constraints}.

Consider a temporal formula $\phi \ra \nexttime \psi$ where 
$\phi$ uses only `past' time operators and
$\psi$ uses only `future' time operators.
Given a  CSP $\setUnion_{i = s}^t \PP_i$, we show how
the past temporal logic formula 
$\phi$ translates to a constraint $\conscd-{\phi}st$
and how
the future temporal logic formula   $\psi$
translates to a constraint $\conscd+{\psi}st$,
both on the variables of $\setUnion_{i = s}^{t} \PP_i$.

We assume that the target constraint language has
Boolean constraints and \emph{reified} versions
of simple comparison and arithmetic constraints.
Reifying a constraint means
associating a Boolean variable with it
that reflects the truth of the constraint.
For example, $(x = y) \equiv b$ is a reified
equality constraint: $b$ is a Boolean variable
reflecting the truth of the constraint $x=y$.

We denote by $\consr{\phi} stb$
the sequence of constraints representing the fact that
the formula $\phi$ has the truth value $b$
in the interval $[s..t]$.
The `past' or `future' aspect of a formula is indicated
by a marker $^-$ or $^+$, resp., when relevant.
The translation of $\phi$ proceeds by induction and is
initiated with $\consr{\phi} st1$
(where $s \leq t$).

%---------------------------------------------------------------------

\subsection{Unfolding Translation}

We translate the propositional connectives
into appropriate Boolean constraints.
The temporal operators are
unfolded over the simulation stages.

For example, the `future' formula
$\eventually (Q[\mt{A}, \mt{B}] = q)$
in the interval $[1..3]$ translates to
\[\begin{array}{l}
	(Q[\mt{A}, \mt{B}, 1] = q) \ \equiv \ b_1,\\
	(Q[\mt{A}, \mt{B}, 2] = q) \ \equiv \ b_2,\\
	(Q[\mt{A}, \mt{B}, 3] = q) \ \equiv \ b_3,\quad\text{and}\\
	b_1 \lor b_2 \lor b_3 = 1,
\end{array}\]
with fresh Boolean variables $b_1,b_2,b_3$.

\paragraph{Translation for `future' formulas.}

\[\begin{array}{l@{\hspace{1em}}l@{\hspace{1.3em}}l}
	\consrd+{\ms{true}} stb
		&\text{is}&
		b = 1;			
		\\
	\consrd+{\lnot\phi} stb
		&\text{is}&
		b' = \lnot b,
		\\&&
		\consrd+{\phi} st{b'};
		\\
	\consrd+{\phi_1 \lor \phi_2} stb
		&\text{is}&
		(b_1 \lor b_2) \equiv b,
		\\&&
		\consrd+{\phi_1} st{b_1},
		\\&&
		\consrd+{\phi_2} st{b_2};
		\\[1ex]
	\consrd+{Q[A,B] \mathbin{?} c} stb
		\quad\text{is}\hspace{-4em}
		\\&&
		\hspace{-7em}
		(Q[A,B,s] \mathbin{?} c)\equiv b
		\text{ where } ? \in \{=,\neq\};
		\\[1ex]
	\consrd+{\nexttime\phi} stb
		&\text{is}&
		(b_1 \land b_2) \equiv b,
		\\&&
		\consrd+{\phi} rt{b_2},
		\\&&
		(s+1 \leq t) \equiv b_1,
		\\&&
		(s+1 = r) \equiv b_1;
		\\[1ex]
	\consrd+{\always\phi} stb
		&\text{is}&
		(\lAnd_{r \in s..t} b_r) \equiv b,
		\\&&
		\hspace{-7em}
		\consrd+{\phi} rt{b_r}
		\text{ for all}\ r \in [s..t];
		\\[1ex]
	\consrd+{\eventually\phi} stb
		&\text{is}&
		(\lOr_{r \in s..t} b_r) \equiv b,
		\\&&
		\hspace{-7em}
		\consrd+{\phi} rt{b_r}
		\text{ for all}\ r \in [s..t];
		\\[1ex]
	\consrd+{\chi \until \phi} stb
		&\text{is}
		\\&&
		\hspace{-7em}
		\consrd+{\phi \lor \chi \land \nexttime(\chi \until \phi)} stb.
\end{array}\]

%---------------------------------------------------------------------

\paragraph{Translation for `past' formulas.}

This case is symmetric to the `future' case
except for the `backward' perspective.
So we have
\[\begin{array}{l}
	\consrd-{Q[A,B] \mathbin{?} c} stb
		\quad\text{is}\hfill
		\\\hspace{7em}
		(Q[A,B,t] \mathbin{?} c) \equiv b
		\text{ where } ? \in \{=,\neq\},
\end{array}\]
for example.
The remaining cases are defined analogously.

\bigskip
Observe that the interval bounds $s,t$
in $\consc{\phi} st$
are treated as constants such that $s \leq t$.

%---------------------------------------------------------------------

\subsection{Array Translation}

This alternative translation avoids the potentially
large disjunctive constraints
caused by unfolding
the $\eventually$ and $\until$ operators.
The idea is to push disjunctive information
inside variable domains,
with the help of \emph{array constraints}.

Reconsider the formula
$\eventually (Q[\mt{A}, \mt{B}] = q)$
in the interval $[1..3]$.
It is translated into a
single array constraint,
with the help of a fresh variable $x$ ranging over
time points:
\[\begin{array}{l}
	Q[\mt{A},\mt{B},x]=q,\\
	1 \leq x, \  x \leq 3.
\end{array}\]

Array constraints generalise
the better-known \textsf{element} constraint.
Constraint propagation for array constraints is studied in
\cite{brand:2001:constraint2} and used in our implementation.

\smallskip
When negation occurs in the formula,
a complication arises with this translation approach,
however.
Just negating the associated truth value,
as in the unfolding translation, is now incorrect.
We therefore first transform a formula into
negation normal form (NNF).

The array translation of NNF formulas follows.
We give it only for `future' formulas
and  where different from the unfolding translation.
The case of negation does not apply anymore.
\[\begin{array}{l@{\quad}ll}
	\consrd+{\always\phi} stb &\text{is}
		\\&&
		\hspace{-6em}
		\consrd+{\phi \land (\nexttime\,\ms{true} \limplies
		\nexttime\always\phi} stb;
		\\[1ex]
	\consrd+{\eventually\phi} stb &\text{is}&
		s \leq r,\ r \leq t,
		\\&&
		\consrd+{\phi} rtb;
		\\[1ex]
	\consrd+{\chi \until \phi} stb &\text{is}&
		(b_1 \land (b_2 \lor b_3)) \equiv b,
		\\&&
		s \leq r,\ r \leq t,
		\\&&
		\consrd+{\phi} rt{b_1},
		\\&&
		(s=r) \equiv b_2,
		\\&&
		s \leq u,\ u \leq r,
		\\&&
		(u = r-1) \equiv b_3,
		\\&&
		\consrd+{\always\chi} su{b_3}.
\end{array}\]

The interval end points $s,t$ in $\consc{\phi} st$
can now be variables with domains,
in contrast to the case of the unfolding translation
where $s,t$ are constants.
We are careful to
maintain the invariant $s \leq t$
and state appropriate constraints to this end.
Therefore, for example, we unfold $\always\phi$ 
into a conjunction only step-wise,
as the formula
$\phi \land (\nexttime\,\ms{true}
 \limplies \nexttime\always\phi)$.

\begin{example}
Let us contrast the two alternative translations
for a formula from the navigation domain.
Consider
\[
	\phi \equiv \eventually( \phi_1 \ \land\ \eventually \phi_2),
\]
$\phi_1 \equiv (Q[\sh,\bu] = \east)$ and
$\phi_2 \equiv (Q[\sh,\bu] = \south)$,
in the interval $[1..n]$
for a constant~$n$,
as a `future' formula.
So we consider the sequence of constraints
$\conscd+{\phi}1n$ for each translation.

The unfolding translation generates many
reified equality constraints
of the form $(Q[\sh,\bu,k] = D)\equiv b_{i,k}$,
where $D$ is $\east$ or $\south$.
More specifically,
$n+\sum_{i=1}^n i = n(n+3)/2$
such constraints and as many
new Boolean variables are created.
Many of the constraints
are variants of each other differing only
in their Boolean variable $b_{i,k}$.

The array translation results in just two array
constraints, namely
$Q[\sh,\bu,r_1] = \east$ and
$Q[\sh,\bu,r_2] = \south$,
The four ordering constraints
$1 \leq r_1$,  $r_1 \leq n$
$r_1 \leq r_2$, and $r_2 \leq n$
control the fresh variables $r_1, r_2$.
\qedex
\end{example}

%=====================================================================

\section{Simulations}\label{sec:simulations}

By a \emph{qualitative simulation} we mean a finite
or infinite sequence $\PS = \tuple{\PP_0, \PP_1, \ldots}$
of CSPs such that for each chosen inter-state constraint
\mbox{$\phi \ra \nexttime \psi$} we have that the constraint
\[
	\cons([0 \mathbin{..} t_0], \phi) \ra \cons([t_0 +1 \mathbin{..} t], \psi)
\]
is satisfied by the CSP $\setUnion_{i = 0}^{t} \PP_{i}$,
\begin{itemize}
\item if $\PS$ is finite with $u$ elements,
	for all
	$t_0 \in [0 \mathbin{..} u-1]$,
	$t = t_{\max}$,
\item if $\PS$ is infinite,
	for all $t_0 \geq 0, t \geq t_0 +1$.
\end{itemize}
Thus, at each stage of the qualitative simulation,
we relate its past (and presence)
to its future using the chosen inter-state constraints.

Consider an initial situation $\I = \PP_0$ and a
final situation $\F_x$ determined by a qualitative array
of the form $Q_x$, where $x$ is a variable ranging
over the set of integers (possible time instances).
We would like to determine whether a simulation exists
that starts in $\I$ and reaches $\F_t$,
where $t$ is the number of steps.
If one exists, we may also be interested in computing
a shortest one, or in computing all of them.

%---------------------------------------------------------------------

\paragraph{Simulation algorithm.}

\setnewlength{\algotab}{2.8em}
\setnewlength{\algotabinitial}{1em}
\setnewlength{\algotabmargin}{1em}
\setnewlength{\algogaplen}{1ex}

\newcommand{\algo}[1]		{\ensuremath{\mathsf{#1}}}
\newcommand{\algosig}[3]	{\ensuremath{\boxed{\mathsf{#1}}:\ {#2}\ \longmapsto\ {#3}}}
\newcommand{\algokw}[1]		{\mbox{\sffamily\textbf{#1}}}
\newcommand{\algogkw}[1]	{\hspace{\algogaplen}\algokw{#1}}
\newcommand{\algokwg}[1]	{\algokw{#1}\hspace{\algogaplen}}
\newcommand{\algogkwg}[1]	{\hspace{\algogaplen}\algokw{#1}\hspace{\algogaplen}}

\newenvironment{algorithm}
	{\begin{tabbing}%
	\hspace{\algotabmargin}\=%
	\hspace{\algotabinitial}\=%
	\hspace{\algotab}\=\hspace{\algotab}\=\hspace{\algotab}\=\hspace{\algotab}\=%
	\hspace{\algotab}\=\hspace{\algotab}\=\hspace{\algotab}\=\hspace{\algotab}\=%
	\kill\+\>}%
	{\end{tabbing}}

\begin{figure}
\begin{algorithm}
\boxed{\algo{Simulate}}\\[1ex]\quad
spatial array $Q$, state constraints, $t_{\max}$
$\longmapsto$ solution
\+\\[1ex]
$\PS := \tuple{}$;\
$t := 0$\\
\algokwg{while} $t < t_{\max}$ \algogkw{do}\+\\
	$\PP_t$ := \algo{create} CSP from $Q_t$ and
	\\\hspace{2.2em}
	impose intra-state constraints\\
	$\PS := $ \algo{append} $\PP_t$ to $\PS$ and
	\\\hspace{2.8em}
	impose inter-state constraints\\[0.7ex]
	$\tuple{\PS, \mi{failure}} := \algo{prop}(\PS)$\\
	\algokwg{if not} $\mi{failure}$ \algogkw{then}\+\\
		$\PS' := \PS$ with final state constraints
		\\\hspace{3.2em}
		imposed on $\PP_t$\\[0.7ex]
		$\tuple{\mi{solution}, \mi{success}} :=
		\algo{solve}(\PS'$)\\
		\algokwg{if} $\mi{success}$ \algogkwg{then return} $\mi{solution}$\-\\
	$t := t+1$\-\\
\algokwg{return failure}\\[-5ex]
\end{algorithm}
\caption{The simulation algorithm}
\label{fig:algorithm}
\end{figure}

The algorithm given in Figure~\ref{fig:algorithm}
provides a solution to the first two problems
in presence of a non-circularity constraint.

The sequence $\PS$ of CSPs is initially empty and
subsequently step-wise extended;
so it remains finite.
We view $\PS$ as a single CSP,
which consists of regular finite domain variables
and constraints and which
thus fits into the problem format
solvable by a standard constraint programming techniques.

We employ the auxiliary procedures
\algo{prop} and \algo{solve}.
The call to $\algo{prop}$ performs constraint propagation
of the intra-state and inter-state constraints.
In our implementation,
the hyper-arc consistency notion is used.
As a result, the variable domains
are pruned and less backtracks arise
when \algo{solve} is called.
If the outcome is an inconsistent CSP,
the value $\ms{false}$ is returned in the $\mi{failure}$ flag.

The call to $\algo{solve}$
checks if a solution to the CSP
corresponding to the given sequence of CSPs exists.
If so, a solution and $\ms{true}$ is returned,
otherwise $\tuple{\emptyset, \ms{false}}$.
In our implementation,
$\algo{solve}$
is a standard backtrack search
(based on variable domain splitting)
combined with constraint propagation
as in the $\algo{prop}$ procedure.

We use the constant $t_{\max}$ equal to the number
of different qualitative arrays, \ie,
$t_{\max} = |\OO| \cdot (|\OO|-1) \cdot 2^{|\Q| -1}$.
If the desired simulation exists, the above algorithm
finds a shortest one and outputs it in the
variable $\mi{solution}$.

%=====================================================================

\section{Implementation}

We implemented the simulation algorithm of
Fig.~\ref{fig:algorithm} and both alternative
translations of temporal formulas to constraints
in the \eclipse{} constraint programming system
\cite{wallace:1997:eclipse}.
The total program size is roughly 1500 lines of code.

\subsection{Propagation}

Support for enforcing hyper-arc consistency
for Boolean and many reified constraints,
as well as for extensionally defined constraints
such as $\conv$, $\comp$ and
the conceptual neighbourhood constraint,
is directly available in \eclipse{}
(by its \textsf{fd/ic} and \textsf{propia} libraries).
For array constraints,
we use the \eclipse{} implementation discussed in
\cite{brand:2001:constraint2}.

The availability of these (generic)
implementations of propagation mechanisms
explains why we chose hyper-arc consistency.
We emphasise, however, that in a relation variable model,
constraint propagation is relevant only for efficiency.

\subsection{Search}

We use the basic backtracking algorithm provided
by \eclipse, but we control it with the heuristics
described in the following section.

Various other, advanced search strategies are
available in \eclipse, for example
Limited Discrepancy Search \cite{harvey:1995:limited}.
Although we did not experiment with these techniques,
we believe it is worth doing so,
and it is not difficult to modify our implementation
(the \algo{solve} procedure) accordingly.

\subsection{Heuristics}

Our implementation also incorporates
the specialised reasoning techniques
for $\RCC$ \cite{renz:2001:efficient}
and the cardinal directions \cite{Lig98}.
In these studies, maximal tractable subclasses
of the respective calculi are identified,
and corresponding polynomial decision procedures
are discussed.

Our context requires that these techniques
are treated as heuristics,
due to the presence of side constraints
(notably the inter-state constraints).
With a relation variable model for qualitative spatial
reasoning, these heuristics fall into the customary class
of variable and value ordering heuristics
for guiding search in constraint programming.

In our implementation,
the search heuristic splits
the relation variable domains appropriately
so that one of the new domains belongs to a
maximal tractable subclass
of the respective calculus.

%=====================================================================

\section{Case Studies}

We now report on two case studies.
In both of them, the solutions were found
by our implementation within a few seconds.

\subsection{Piano Movers Problem}

Consider the following version of the piano movers problem.
There are three rooms,
the living room (\texttt{L}),
the study room (\texttt{S}) and
the bedroom (\texttt{B}),
and the corridor (\texttt{C}).
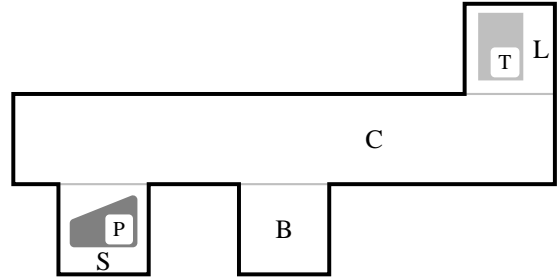
\begin{figure}
\centering
\psset{unit=6mm}
\begin{pspicture}(0,0)(12,6)
\psline[linecolor=lightgray](1,2)(3,2)
\psline[linecolor=lightgray](5,2)(7,2)
\psline[linecolor=lightgray](10,4)(12,4)
\pspolygon[linewidth=1.5pt](0,2)(1,2)(1,0)(3,0)(3,2)(5,2)(5,0)(7,0)(7,2)(12,2)(12,6)(10,6)(10,4)(0,4)
\pspolygon*[linecolor=gray,linearc=0.1](1.25,0.6)(2.75,0.6)(2.75,1.8)(1.25,1.2)
\rput(2.35,1){\psframebox*[framearc=0.3]{{\footnotesize P}}}
\psframe*[linecolor=lightgray](10.3,4.3)(11.3,5.8)
\rput(10.9,4.7){\psframebox*[framearc=0.3]{{\footnotesize T}}}
\rput(6,1){B}
\rput(8,3){C}
\rput(2,0.3){S}
\rput(11.7,5){L}
\end{pspicture}
\caption{A piano movers problem}
\label{fig:piano}
\end{figure}
Inside the study room there is a piano (\texttt{P}) and inside the living
room a table (\texttt{T}); see Figure \ref{fig:piano}.
Move the piano to the living room and the table
to the study room assuming that none of the rooms and the corridor are large enough to
contain the piano and the table at the same time.
Additionally, ensure that the piano and the table at no time will touch each other.
%\end{quote}

To formalise this problem,
we describe the initial situation
by means of the following formulas:
\[\begin{array}{@{}l@{\hspace{0.8em}}l@{\hspace{1em}}l}
	\phi_0 &\equiv&
		Q[\texttt{B,L}] = \disjoint \ \land \
		\\&&
		Q[\texttt{B,S}] = \disjoint \ \land \
		\\&&
		Q[\texttt{L,S}] = \disjoint,\\[1ex]
	\phi_1 &\equiv&
		Q[\texttt{C,B}] = \meet \ \land \
		\\&&
		Q[\texttt{C,L}] = \meet \ \land \
		\\&&
		Q[\texttt{C,S}] = \meet,\\[1ex]
	\phi_2 &\equiv&
		Q[\texttt{P,S}] = \inside \ \land \
		\\&&
		Q[\texttt{T,L}] = \inside.
\end{array}\]
We assume that initially $\phi_0$, $\phi_1$, $\phi_2$ hold,
\ie, the constraints
$\cons^{-}([0\mathbin{..}0], \phi_0)$,
$\cons^{-}([0\mathbin{..}0], \phi_1)$ and
$\cons^{-}([0\mathbin{..}0], \phi_2)$
are present in the initial situation $\I$.

Below, given a formula $\phi$, by an \emph{invariant built out of} $\phi$
we mean the formula $\phi \ra \nexttime \always \phi$.
Further, we call a room or a corridor a `space' and abbreviate the
subset of objects $\{ \mt{B},\mt{C},\mt{L},\mt{S} \}$ by $\SSS$.
We now stipulate as the inter-state constraints
the invariants built out of the following formulas:
\begin{itemize}
\item the relations between the rooms,
	and between the rooms and the corridor, do not change:
	$\phi_0 \land \phi_1$,
\item at no time do the piano and the table fill
	completely any space:
	\[
	\forall s \in \SSS.\, \left(
			Q[\mt{P}, s] \neq \equal
			\ \land \
			Q[\mt{T}, s] \neq \equal
			\right),
	\]
\item together, the piano and the table do not fit into any space.
	More precisely, at each time, at most one of these two objects
	can be within any space:
	\[\begin{array}{l}
	\forall s \in \SSS.\,
		\neg
		(
		Q[\mt{P},s] \in \{\inside, \coveredby\}
		{}\land{}
		\\\hspace{4.4em}
		Q[\mt{T},s] \in \{\inside, \coveredby\}
		),
	\end{array}\]
\item at no time instance do the piano and
	the table touch each other:
	\[
		Q[\mt{P},\mt{T}] = \disjoint.
	\]
\end{itemize}
The final situation is captured by the constraints
\[\begin{array}{l}
	Q[\mt{P,L}] = \inside
	\quad\text{and}\quad
	Q[\mt{T,S}] = \inside.
\end{array}\]

Remarkably, the interaction with our program revealed
in the first place that
our initial formalisation was incomplete.
For example, the program also generated solutions
in which the piano is
moved not through the corridor
but `through the walls', as it were.

To avoid such solutions we added 
the following intra-state constraints.
\begin{itemize}
\item each space is too small to be `touched' (\emph{met})
	or `overlapped' by the piano and the table
	at the same time:
	\[\begin{array}{l}
	\forall s \in \SSS.\
		\neg (Q[s,\mt{P}] \in \{\overlap, \meet\}
		{}\land{}
		\\\hspace{4.4em}
		Q[s,\mt{T}] \in \{\overlap, \meet\}),
	\end{array}\]
\item if the piano or the table overlaps with one space $s$,
	then it also overlaps with some other space $s'$,
	such that $s$ and $s'$ \emph{touch} each other:
	\[\begin{array}{l}
		\forall s \in \SSS.\,
		\forall o \in \{\mt{P},\mt{T}\}.\,
		(Q[s,o] = \overlap \raa
		\\\hspace{1.2em}
		\exists s' \in \SSS.
		\left(Q[s',o] = \overlap \land Q[s,s'] = \meet\right)),
	\end{array}\]
\item if the piano overlaps with one space, then
	it does not \emph{touch} any space,
	and equally the table:
	\[\begin{array}{l}
	\forall s \in \SSS.\
	 \forall o \in \{\mt{P},\mt{T}\}.
	\\\hspace{1.5em}(Q[s,o] = \overlap \ra
	\forall s' \in \SSS.\ Q[s',o] \neq \meet),
	\end{array}\]
\item both the piano and the table can
	\emph{touch} at most one space at a time:\\
	\[\begin{array}{l}
	       \forall s, s' \in \SSS.\
		\forall o \in \{\mt{P},\mt{T}\}.
		\\\hspace{1.5em}
		(Q[s,o] = \meet \land Q[s',o] = \meet
		\raa{}
		\\\hspace{13em}
		Q[s,s'] = \equal).
	\end{array}\]
\end{itemize}

After these additions, our program generated the shortest
solution in the form of a simulation of length 12.
In this solution the bedroom is used as a temporary storage
for the table.  Interestingly, the table is not moved
completely into the bedroom: at a certain moment it only
overlaps with the bedroom.

%---------------------------------------------------------------------

\subsection{Phagocytosis}

The second example deals with a simulation of phagocytosis:
an amoeba absorbing a food particle.
This problem is discussed in \cite{CCR92a}.
We quote:

\begin{quote}
  ``Each amoeba is credited with vacuoles (being fluid spaces)
  containing either enzymes or food which the animal has digested. The
  enzymes are used by the amoeba to break down the food into nutrient
  and waste. This is done by routing the enzymes to the food vacuole.
  Upon contact the enzyme and food vacuoles fuse together and the
  enzymes merge into the fluid containing the food. After breaking
  down the food into nutrient and waste, the nutrient is absorbed into
  the amoeba's protoplasm, leaving the waste material in the vacuole
  ready to be expelled.  The waste vacuole passes to the exterior of
  the protozoan's (\ie, amoeba's) body, which opens up, letting the
  waste material pass out of the amoeba and into its environment.''
\end{quote}

To fit it into our present framework, we slightly simplified
the problem representation
by not allowing for objects to be added or removed dynamically. %during the simulation.

\smallskip
In this problem, we have six objects,
\so{amoeba}, \so{nucleus}, \so{enzyme}, \so{vacuole},
\so{nutrient} and \so{waste}.
The initial situation is described by means
of the three following constraints:
\[\begin{array}{l}
	Q[\so{amoeba}, \so{nutrient}] = \disjoint,\\
	Q[\so{amoeba}, \so{waste}] = \disjoint,\\
	Q[\so{nutrient}, \so{waste}] = \equal.\hspace{10em}
\end{array}\]
We have the intra-state constraints
\[\begin{array}{l}
	Q[\so{enzyme}, \so{amoeba}] = \inside,\\
	Q[\so{vacuole}, \so{amoeba}] \in \{\inside, \coveredby\},\\
	Q[\so{vacuole}, \so{enzyme}] \in \{\disjoint,\meet,\overlap,\covers\},
\end{array}\]
and, concerning the nucleus,
\[\begin{array}{l}
	Q[\so{nucleus}, \so{vacuole}] \in \{\disjoint, \meet\},\\
	Q[\so{nucleus}, \so{enzyme}] \in \{\disjoint, \meet\},\\
	Q[\so{nucleus}, \so{amoeba}] = \inside.
\end{array}\]

The inter-state constraints are
\[\begin{array}{l}
	Q[\so{nutrient}, \so{amoeba}] = \meet \raa{}
	\\\hspace{9em}
	\nexttime\ Q[\so{nutrient, amoeba}] = \overlap,
	\\[1ex]
	Q[\so{nutr.}, \so{amoeba}] \in \{\inside, \coveredby, \overlap \raa{}
	\\\hspace{6em}
	\nexttime\ Q[\so{nutr.}, \so{amoeba}] \in \{\inside, \coveredby\}.
\end{array}\]
We model the splitting up of the food
into nutrient and waste material by
\[\begin{array}{l}
	Q[\so{nutrient}, \so{waste}] = \equal
	\ \mathbin{\Dot\raa}{}\\[0.3ex]
	\quad (\phi_1 \mathbin{\Dot\ra} \phi_2 \mathbin{\Dot\lor}\phi_3)
	\\[0.3ex]
	\quad \mathbin{\Dot\lor}{}\\[0.3ex]
	\quad \nexttime \ Q[\so{nutrient}, \so{waste}] \neq \equal;
\end{array}\]
with
\[\begin{array}{lll}
	\phi_1 &\equiv&
	Q[\so{nutrient}, \so{vacuole}] = \inside \land {}
	\\&&	Q[\so{enzyme}, \so{nutrient}] = \overlap \land {}
	\\&&	Q[\so{enzyme}, \so{waste}] = \overlap
	\\[1ex]
	\phi_2 &\equiv&
	\nexttime\ Q[\so{nutrient}, \so{waste}] = \overlap
	\\[1ex]
	\phi_3 &\equiv&
	\nexttime\ Q[\so{nutrient}, \so{waste}] = \equal\\
\end{array}\]
The dotted operators express \emph{if-then-else},
that is,
	\[
	a \mathbin{\Dot\ra} b \mathbin{\Dot\lor} c
	\quad\equiv\quad
	(a \ra b) \land (\neg a \ra c).
	\]

The final situation is described by means of the constraints
\[\begin{array}{l}
	Q[\so{amoeba}, \so{waste}] = \disjoint,\\[0.2ex]
	Q[\so{amoeba}, \so{nutrient}] \in \{\contains, \covers\}.
\end{array}\]
Our program generated a simulation consisting of 9 steps.

%=====================================================================

\section{Final Remarks}

The most common approach to qualitative simulation is
the one discussed in \cite[chapter 5]{Kui94}.
For a recent overview see \cite{Kui01}.
It is based on a qualitative differential equation model
(QDE) in which one abstracts from the usual differential
equations by reasoning about a finite set of symbolic values
(called \emph{landmark values}).
The resulting algorithm, called \emph{QSIM}, constructs
the tree of possible evolutions by repeatedly constructing the
successor states. During this process, CSPs are generated and solved.

This approach is best suited to simulate evolution of
physical systems.  A standard example is a simulation of
the behaviour of a bath tub with an open drain and constant
input flow.  The resulting constraints are usually
equations between the relevant variables and lend
themselves naturally to a formalisation using CLP(FD), see
\cite[chapter 20]{bratko:2001:prolog} and
\cite{bandelj:2002:qualitative}.
The limited expressiveness of this approach was overcome in
\cite{brajnik:1998:focusing}, where branching time temporal
logic was used to describe the relevant constraints on the
possible evolutions (called `trajectories' there).  This
leads to a modified version of the \emph{QSIM} algorithm in
which model checking is repeatedly used.

Our approach is inspired by the qualitative spatial
simulation studied in \cite{CCR92a}, the main features of
which are captured by the composition table and the
neighbourhood relation discussed in Example \ref{ex:rcc8}.
The distinction between the intra-state and inter-state
constraints is introduced there, however the latter only
link the consecutive states in the simulation. The
simulation algorithm of
\cite{CCR92a} generates a complete tree of all `evolutions',
usually called an \emph{envisionment}.

In contrast to \cite{CCR92a}, our approach is
constraint-based.
This allows us to repeatedly use constraint propagation to
prune the search space in the simulation algorithm.
Further, by using more complex inter-state constraints,
defined by means of temporal logic, we can express
substantially more sophisticated forms of behaviour.

While the prevalent approach to constraint-based
modelling of qualitative spatial knowledge
maps qualitative relations to constraints,
we use variables to express qualitative relations.
The relation variable approach is much more declarative,
separating the model from the solver.
The advantage of a relation variable model
for qualitative simulations is that the knowledge
of the spatial domain as well as of the application domain
can be expressed on the same conceptual level,
by intra-state and inter-state constraints.
This leads to a model that can easily be realised
within a typical constraint programming system
using generic propagation and search techniques,
and is also immediately open to advances in these systems.

Simulation in our approach subsumes a form of planning. In
this context, we mention the related work
\cite{lopez:2003:generalizing} in the area of planning
which shows the benefits of encoding planning problems as
CSPs and the potential with respect to solving efficiency.
Also related is the \textsc{TLplan} system where planning
domain knowledge is described in temporal logic
\cite{bacchus:2000:using}. The planning system is based on
incremental forward-search, so temporal formulas are just
unfolded one step at a time, in contrast to the translation
into constraints in our constraint-based system.

Finally, \cite{Fal00} discusses how a qualitative version
of the piano movers problem can be solved using an approach
to qualitative reasoning based on topological inference and
graph-theoretic algorithms. Our approach is substantially
simpler in that it does not rely on any results on topology
apart of a justification of the composition table.

%---------------------------------------------------------------------

\bibliographystyle{plain}
\bibliography{bibfile}

\end{document}